# Multiple Imputation for Biomedical Data using Monte Carlo Dropout Autoencoders

Kristian Miok[1], Dong Nguyen-Doan[1], Marko Robnik-Šikonja[2], Daniela Zaharie[1]
[1]Department of Computer Science, Faculty of Mathematics and Computer Science, West University of Timisoara, Timisoara, Romania, {kristian.miok, dong.nguyen10, daniela.zaharie}@e-uvt.ro
[2]Faculty of Computer and Information Science, University of Ljubljana, Ljubljana, Slovenia, marko.robnik@fri.uni-lj.si

*Abstract*—Due to complex experimental settings, missing values are common in biomedical data. To handle this issue, many methods have been proposed, from ignoring incomplete instances to various data imputation approaches. With the recent rise of deep neural networks, the field of missing data imputation has oriented towards modelling of the data distribution. This paper presents an approach based on Monte Carlo dropout within (Variational) Autoencoders which offers not only very good adaptation to the distribution of the data but also allows generation of new data, adapted to each specific instance. The evaluation shows that the imputation error and predictive similarity can be improved with the proposed approach.

## I. INTRODUCTION

Missing values are common in data science applications. Incomplete data usually lead to a decrease in models' quality and even to wrong insights [1]. In medical framework, where decisions based on incomplete data can have important consequences, the simplest solution of dropping instances with missing values is unacceptable. Similarly, in biomedical applications, there is often not enough data and existing instances are difficult to obtain. On the other hand, data imputation techniques replace missing values with substitutes [2]. However, reasons for missing data need to be clarified before imputation. There are three main types of missing data: *missing completely at random* (MCAR), *missing at random* (MAR), and *missing not at random* (MNAR). Values are considered MCAR when the probability that they are missing is independent both with the respect to possible values of the variable and with respect to other observable variables in the data. The MAR case is characterized by the fact that the missing probability can be estimated from existing variables. When missing data is neither MCAR nor MAR, it is MNAR. In this case, reasons for a missing value can depend on other variables as well as on the analyzed variable. In the context of this paper, we assume the MCAR case.

The classical imputation methods generate a single value for one missing observation. The shortcoming of such an approach is that the uncertainty of the imputation process is not taken into account, and the generated value is implicitly considered correct. Actually, this assumes that the imputation model is perfect and does not to account for error in the imputation process. In order to overcome this drawback, each missing value could be replaced with several different values, reflecting the uncertainty about the missing information. This technique is frequently used for missing data imputation and is called *multiple imputation* (MI) [3]. However, due to different underlying assumptions, the challenge of MI face is a correct specification of the imputation model [4]. For instance, some imputation models are incapable of handling mixed data types (categorical and continuous), some have strict distributional assumptions (multivariate normality), and/or cannot handle arbitrary missing data patterns. The classical methods capable of overcoming these problems can be limited with respect to modeling highly nonlinear relationships, large amounts of data, and complicated attribute interactions that need to be preserved.

Recent advances in deep learning have established state-of-the-art results in many fields [5]. Deep architectures have the capability to automatically learn latent representations and complex inter-variable associations. Supervised and unsupervised methods like autoencoders (AE), variational autoencoders, and Generative Adversarial Networks (GANs) are applied within a range of practical problems. On the other hand, due to a pressing need for better understanding of model reliability, the Bayesian inference is increasingly used for deep learning models. One of the most efficient implementations of the Bayesian uncertainty estimation in neural networks is the Monte Carlo Dropout (MCD) method [6]. We combine unsupervised autoencoder modeling with Bayesian uncertainty estimation into a generative model aimed to improve existing state-of the art results in the missing data imputation.

The paper consists of six sections. In the Section II, we present a short overview of previous work done on the imputation methods. In Section III, we first describe the components forming our approach: autoencoders and variational autoencoders, followed by the main idea of our approach, i.e. the MCD method within autoencoders adapted for data imputation. We use denoising autoencoder methods as unsupervised learning methods and train a deep neural network to reconstruct an input that has been corrupted by missing values. The idea is to generate multiple outcomes for a single missing value and average the results using the MCD method. Section 4 describes the experimental scenario, as well as the publicly available data sets and the motivating real-world data set. Results of the experiments are presented in Section V, and Section VI concludes the paper.

## II. PREVIOUS WORK

The missing data imputation methods have been extensively investigated in many applicative areas, particularly in the biomedical domain. Due to complex experimental setting, limited budget, and participants that can choose to stop the experiments, even controlled studies in (bio)medical field have to deal with this problem. The prominent imputation methods keep the full data set size, but potentially introduce different kinds of imputation biases. A variety of imputation approaches were proposed [7]. Examples of simple approaches of are mean imputation, last value carried forward, and imputation based on logical rules. More sophisticated methods include random forest based imputation, and deep neural network based imputation [8, 9]. Deep generative models are highly flexible and can capture the latent structure of complex high-dimensional data to generate new values [3].

The most popular multiple imputation method among the classical approaches is MICE [10,11]. MICE consist of four different steps and imputes data on a variable by variable basis by specifying the imputation model per variable [12].

Deep generative models proved to be very powerful for various tasks such as computer vision and text mining. Encouraged with such results, researchers from different domains investigated the deep learning application to the generative and missing values imputation tasks. The first deep autoencoder model that performs multiple imputation was proposed by [20]. Other approaches are based on Generative Adversarial Networks (GANs) [13]. For instance, medGAN has been proposed to generate synthetic health care patient records [14], the GAN has been used to generate synthetic passenger name records [15], and Generative Adversarial Imputation Nets (GAIN) impute missing components conditioned on what is actually observed [16]. Different types of variational autoencoders have been proposed as data generators: from the text generating tasks [17] to the tasks where generated data shall preserve the distributional properties of the original data [18]. A method that enables a variational autoencoder to handle incomplete heterogenous data was proposed in [19] while the model introduced by [20] achieves multiple imputation based on overcomplete deep denoising autoencoders.

## III. METHODS

In contrast to imputation techniques that impute only single missing value, in this work, we investigate generative autoencoders that can create a new data set, similar to the original one. The most popular methods used for generating new values from the observed ones within the medical data domain are GANs. GAN models consists of two parts: the generator and the discriminator. The generator brings noise into the generated results and tries to "fool" the discriminator by generating some instances far away from the original data distribution. In case the discriminator fails to perform its job properly, some of the generated instances can be far from the expected data distribution. For this reason, using GANs within the medical domain is questionable. Thus, there is a need for a deep learning generator that can be trusted. Thus, our main focus in this work is on improving existing autoencoder models. In this section, we first provide a brief overview of autoencoders and variational autoencoders, followed by the incorporation of Monte Carlo dropout within them.

### A. Autoencoders (AEs)

Deep learning models rely on extraction of complex features from data. The goal is to transform the input from its raw format, to another, lower dimensional representation using a sequence of transformations incorporated in several layers of functional units. The resulting representation shall contain features that describe hidden characteristics of the input. Autoencoders (AEs) are deep learning models made of the two parts: an *encoder* network that compresses high-dimensional input data into a lower-dimensional representation vector and a *decoder* network that decompresses a given representation vector back to the original form [21]. As the reconstruction is based on the latent lower-dimensional space, AE can be used for both the dimensionality reduction task and to generate new observations from the distribution of the original data.

### B. Variational Autoencoders (VAEs)

In 2013, Kingma and Welling [22] published a foundation for *Variational Autoencoder* (VAE) neural network, becoming one of the fundamental generative models. The main distinction between VAEs and AEs is the learning process where VAEs explicitly estimate the distribution from which the latent space is sampled [23]. Hence, VAEs store the latent variables in the form of probability distributions. This allows easy random sampling and interpolation.

### C. Monte Carlo Dropout Method for Data Imputation

Since its introduction by Gal and Ghahramani in 2016 [6], the Monte Carlo Dropout (MCD) method has been implemented within various neural networks architectures, like convolutional and recurrent networks [24,25]. The properties of dropout [26] allowed its implementations within different machine learning prediction tasks [27,28,18]. In this work, we apply the MCD within the AE and VAE decoder layers in order to obtain multiple generated inputs. By averaging the output values for a specific missing value, we achieve better imputation accuracy than by using the classical approaches. The implementation of MCD within the AE and VAE (MCD-(V)AE) was first time proposed by [18] with the intention to improve subject specific generation from AE and VAE models. The architecture of this method is described in Fig1.

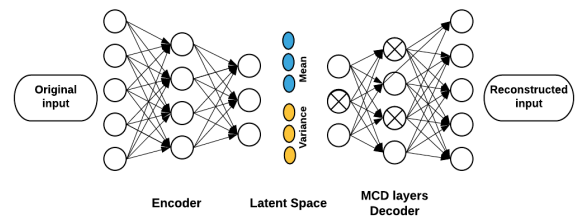

**Figure 1.** Architecture of the Variational Autoencoder with Monte Carlo dropout decoder.

Once the autoencoder models are fitted on the training set, unused instances with missing values are used as an input to generate a new dataset that shall ideally contain the information from the original dataset and the input instances. The idea is to construct a model that absorbs most information from the data and uses dimensionality reduction property of AE or VAE to reduce the noise of random missingness. As shown in [18] the advantage of MCD-(V)AE model over other autoencoders is that it captures the specific distribution of the provided observations from which the generation is done. Hence, not relying on all the data can improve imputation error and better preserve true relationship among the attributes.

## IV. EXPERIMENTS

### A. Data sets

In our experiments, we used medical data from the UCI (University of California Irvine) repository, as well as the data on characteristics of milk collected at a Research and Development Station for Bovine, Arad, Romania in the context of a research project aiming to study the antimicrobial resistance induced in dairy animals. The data collected from 264 cows contain the following attributes: milk quantity, casein, fat, lactose, Ph level, protein, urea, and the number of somatic cells (NCS – an indicator of a potential infection). The characteristics of used datasets are presented in Table 1.

**Table 1:** The characteristics of the used data sets: N – the number of instances, a – the number of attributes, num – the number of numeric attributes, disc – the number of discrete attributes, C – the number of class values.

| Datasets | N | a | num | disc | C |
|---|---|---|---|---|---|
| Study-MILK | 610 | 11 | 11 | 1 | 2 |
| Brest-WISC | 699 | 9 | 9 | 1 | 2 |
| PIMA-diabetes | 768 | 8 | 8 | 1 | 2 |

### B. Experimental Scenario

In our experiment, we only considered the instances with no missing values. To control the amount of missing values, we removed 10%, 30%, and 50% of the values randomly. In this way, missing values can be considered the result of a completely random process (MCAR), and the methods are fairly compared. In the tests, the missing values are masked by setting them all to -1. Each dataset is split into a training set (90%), on which the models are trained, and a testing set (10%), which is used for measuring the performance.

Generative models, which produce new datasets, are more difficult to evaluate. We used two measures: the imputation error measured with the root mean squared error (RMSE), and the predictive performance measured with the classification accuracy. The ability to preserve the prediction properties of the original data set was evaluated using the framework proposed by [29]. The idea is to compare the predictive properties of the two datasets, original one and the newly generated one, by computing the difference between the accuracies of the two corresponding models $\Delta acc =$ accuracy(dataset1) – accuracy(dataset2). If this difference is close to 0, the new dataset preserves most of the predictive performance of the original dataset.

### C. Implementation Details

The architecture of all the encoders includes 2 hidden layers with the size of 80 and 20, and the dropout rate of 0.2. With the MCD method, we generate multiple predicted values, and calculate the mean values of the results. We train all the models for 300 epochs. The link to the source code is provided at https://github.com/KristianMiok/MI_MCD_VAE.

## V. RESULTS

We use the RMSE for the imputation error and $\Delta acc$ for the predictive performance. The imputation error results for the four imputation models using 5-fold cross-validation are presents in the Tables 2, 3 and 4. The lowest error for each dataset is set in bold typeface.

**Table 2:** Comparing imputation error using RMSE for 10% missing data.

| 10% missing data (RMSE) | | | |
|---|---|---|---|
| **Models** | **MILK** | **WISC** | **PIMA** |
| **AE** | 0.05364[0.0011] | 0.07649[0.0093] | 0.06565[0.0059] |
| **VAE** | 0.05027[0.0015] | 0.06014[0.0038] | 0.06909[0.0083] |
| **MCD-AE** | 0.0479 [0.0015] | 0.06048[0.0053] | 0.06649[0.0082] |
| **MCD-VAE** | **0.0465 [0.0012]** | **0.05939[0.0029]** | **0.06462[0.0088]** |

**Table 3:** Comparing imputation error using RMSE for 30% missing data.

| 30% missing data (RMSE) | | | |
|---|---|---|---|
| **Models** | **MILK** | **WISC** | **PIMA** |
| **AE** | 0.09755[0.0080] | 0.12444[0.0170] | **0.11103[0.0046]** |
| **VAE** | 0.08049[0.0049] | 0.11229[0.0091] | 0.11666[0.0054] |
| **MCD-AE** | 0.08685[0.0049] | 0.1129 [0.0105] | 0.11410[0.0048] |
| **MCD-VAE** | **0.07827[0.0051]** | **0.1059 [0.0080]** | 0.11221[0.0051] |

**Table 4:** Comparing imputation error using RMSE for 50% missing data.

| 50% missing data (RMSE) | | | |
|---|---|---|---|
| **Models** | **MILK** | **WISC** | **PIMA** |
| **AE** | 0.12491[0.0104] | 0.14901[0.0216] | 0.14132[0.0108] |
| **VAE** | 0.10002[0.0050] | 0.13753[0.0092] | 0.14057[0.0072] |
| **MCD-AE** | 0.10559[0.0052] | **0.12488[0.0091]** | 0.13829[0.0074] |
| **MCD-VAE** | **0.09764 [0.0053]** | 0.12706[0.0128] | **0.13815[0.0072]** |

The RMSE values in the three tables suggest that the imputation error is reduced when MCD is used. While there is no clear pattern how the imputation error change with the percentage of missing values, it seems the MCD methods work well for all the three proportions of missing data. Apart for the PIMA dataset with 30% of missing values where AE model provides the best results, the MCD models are the preferred imputation methods in all the three experimental datasets.

**Table 5:** Comparing predictive properties between original and imputed data using $\Delta acc$ score for 10% of missing values.

| Models | MILK | WISC | PIMA |
|---|---|---|---|
| **AE** | -0.0131 | 0.01428 | 0.1365 |
| **VAE** | -0.0172 | -0.00714 | 0.1429 |
| **MCD-AE** | **-0.009** | **0** | **0.1230** |
| **MCD-VAE** | -0.010 | **0** | 0.1492 |

The differences in the prediction score Δacc is presented in Table 5 for 10 % of missing value. The results show that using MCD within the AE and VAE models preserve relationships among the variables present in the original data. The low Δacc values for MCD methods indicate that prediction properties of original and generated data are similar.

## VI. Conclusions

We investigate the Monte Carlo Dropout method within autoencoders and variational autoencoders used for multiple imputation. The obtained results, measured with the imputation error and difference in prediction accuracy, suggest that the data imputation performance of autoencoders is improved by using the MCD method.

While most of the imputation methods assume the data is missing at random, in the practice this is not always the case. Our further work will focus on the deep learning models that can handle more specific missing data problems.

## Acknowledgment

The research was carried out in the frame of the project Bioeconomic approach to antimicrobial agents – use and resistance financed by UEFISCDI by contract no. 7PCCDI/2018, cod PN-III-P1-1.2-PCCDI-2017-0361 (Kristian Miok and Daniela Zaharie). Also, the work was supported by the Slovenian Research Agency (ARRS) core research programme P6-0411 (Marko Robnik-Šikonja) and received funding from the European Unions Horizon 2020 research and innovation programme under grant agreement No 825153 (EMBEDDIA) (Kristian Miok and Marko Robnik-Šikonja).

## References

[1] Lall, R. (2016). How multiple imputation makes a difference. *Political Analysis*, *24*(4), 414-433.

[2] Camino, R. D., Hammerschmidt, C. A., & State, R. (2019). Improving Missing Data Imputation with Deep Generative Models. *arXiv preprint arXiv:1902.10666*.

[3] Little, R. J., & Rubin, D. B. (2019). *Statistical analysis with missing data* (Vol. 793). John Wiley & Sons.

[4] Morris, T. P., White, I. R., & Royston, P. (2014). Tuning multiple imputation by predictive mean matching and local residual draws. *BMC medical research methodology*, *14*(1), 75.

[5] LeCun, Y., Bengio, Y., & Hinton, G. (2015). Deep learning. *nature*, *521*(7553), 436.

[6] Gal, Y., & Ghahramani, Z. (2016, June). Dropout as a bayesian approximation: Representing model uncertainty in deep learning. In *international conference on machine learning* (pp. 1050-1059).

[7] Van Buuren, S. (2018). *Flexible imputation of missing data*. Chapman and Hall/CRC.

[8] Tang, F., & Ishwaran, H. (2017). Random forest missing data algorithms. *Statistical Analysis and Data Mining: The ASA Data Science Journal*, *10*(6), 363-377.

[9] Gheyas, I. A., & Smith, L. S. (2010). A neural network-based framework for the reconstruction of incomplete data sets. *Neurocomputing*, *73*(16-18), 3039-3065.

[10] Raghunathan, T. E., Lepkowski, J. M., Van Hoewyk, J., & Solenberger, P. (2001). A multivariate technique for multiply imputing missing values using a sequence of regression models. *Survey methodology*, *27*(1), 85-96.

[11] Van Buuren, S. (2007). Multiple imputation of discrete and continuous data by fully conditional specification. *Statistical methods in medical research*, *16*(3), 219-242.

[12] Azur, M. J., Stuart, E. A., Frangakis, C., & Leaf, P. J. (2011). Multiple imputation by chained equations: what is it and how does it work?. *International journal of methods in psychiatric research*, *20*(1), 40-49.

[13] Goodfellow, I., Pouget-Abadie, J., Mirza, M., Xu, B., Warde-Farley, D., Ozair, S., & Bengio, Y. (2014). Generative adversarial nets. In *Advances in neural information processing systems* (pp. 2672-2680).

[14] Choi, E., Biswal, S., Malin, B., Duke, J., Stewart, W. F., & Sun, J. (2017). Generating multi-label discrete patient records using generative adversarial networks. *arXiv preprint arXiv:1703.06490*.

[15] Mottini, A., Lheritier, A., & Acuna-Agost, R. (2018). Airline passenger name record generation using generative adversarial networks. *arXiv preprint arXiv:1807.06657*.

[16] Yoon, J., Jordon, J., & Van Der Schaar, M. (2018). Gain: Missing data imputation using generative adversarial nets. *arXiv preprint arXiv:1806.02920*.

[17] Wang, W., Gan, Z., Xu, H., Zhang, R., Wang, G., Shen, D., & Carin, L. (2019). Topic-Guided Variational Autoencoders for Text Generation. *arXiv preprint arXiv:1903.07137*.

[18] Miok, K., Nguyen-Doan, D., Zaharie, D., & Robnik-Šikonja, M. (2019). Generating Data using Monte Carlo Dropout. *arXiv preprint arXiv:1909.05755*.

[19] Nazabal, A., Olmos, P. M., Ghahramani, Z., & Valera, I. (2018). Handling incomplete heterogeneous data using vaes. *arXiv preprint arXiv:1807.03653*.

[20] Gondara, L., & Wang, K. (2018, June). Mida: Multiple imputation using denoising autoencoders. In *Pacific-Asia Conference on Knowledge Discovery and Data Mining* (pp. 260-272). Springer, Cham.

[21] Rumelhart, D. E., Hinton, G. E., & Williams, R. J. (1985). *Learning internal representations by error propagation* (No. ICS-8506). California Univ San Diego La Jolla Inst for Cognitive Science.

[22] Kingma, D. P., & Welling, M. (2013). Auto-encoding variational bayes. *arXiv preprint arXiv:1312.6114*.

[23] Doersch, C. (2016). Tutorial on variational autoencoders. *arXiv preprint arXiv:1606.05908*.

[24] Gal, Y., & Ghahramani, Z. (2015). Bayesian convolutional neural networks with Bernoulli approximate variational inference. *arXiv preprint arXiv:1506.02158*.

[25] Gal, Y., & Ghahramani, Z. (2016). A theoretically grounded application of dropout in recurrent neural networks. In *Advances in neural information processing systems* (pp. 1019-1027).

[26] Helmbold, D. P., & Long, P. M. (2017). Surprising properties of dropout in deep networks. *The Journal of Machine Learning Research*, *18*(1), 7284-7311.

[27] Miok, K. (2018). Estimation of Prediction Intervals in Neural Network-Based Regression Models. In *2018 20th International Symposium on Symbolic and Numeric Algorithms for Scientific Computing (SYNASC)* (pp. 463-468). IEEE.

[28] Miok, K., Nguyen-Doan, D., Škrlj, B., Zaharie, D., & Robnik-Šikonja, M. (2019, October). Prediction Uncertainty Estimation for Hate Speech Classification. In *International Conference on Statistical Language and Speech Processing* (pp. 286-298). Springer, Cham.

[29] Robnik-Šikonja, M. (2018). Dataset comparison workflows. *International Journal of Data Science*, *3*(2), 126-14